\documentclass{article} 
\usepackage{preprint_iclr2020, times}
\iclrfinalcopy

\usepackage{amsmath,amsfonts,bm}

\def\eqref#1{~(\ref{#1})}
\def\Eqref#1{Equation~(\ref{#1})}

\def\1{\bm{1}}

\DeclareMathAlphabet{\mathsfit}{\encodingdefault}{\sfdefault}{m}{sl}
\SetMathAlphabet{\mathsfit}{bold}{\encodingdefault}{\sfdefault}{bx}{n}

\def\gB{{\mathcal{B}}}

\def\gD{{\mathcal{D}}}

\def\gM{{\mathcal{M}}}

\def\gO{{\mathcal{O}}}

\newcommand{\E}{\mathbb{E}}

\newcommand{\R}{\mathbb{R}}

\DeclareMathOperator*{\argmin}{arg\,min}

\usepackage{hyperref}
\usepackage{url}
\usepackage{qiangstyle}

\usepackage{wrapfig}

\title{Energy-Aware Neural Architecture Optimization with 
 Fast Splitting Steepest Descent}

\author{Dilin Wang\textsuperscript{1},  Meng Li\textsuperscript{2}, Lemeng Wu\textsuperscript{1}, Vikas Chandra\textsuperscript{2}, Qiang Liu \textsuperscript{1} \\
\textsuperscript{1} Department of Computer Science, UT Austin\\
\textsuperscript{2} Facebook Inc. \\
\texttt{\{dilin, lmwu, lqiang\}@cs.utexas.edu, \{meng.li, vchandra\}@fb.com} \\
}

\begin{document}

\maketitle

\begin{abstract}
Designing energy-efficient networks is 
of critical importance for 
enabling state-of-the-art deep learning in mobile and edge settings where the computation and energy budgets are highly limited. 
Recently, \citet{splitting2019} framed the search of efficient neural architectures into a \emph{continuous splitting process}: it iteratively splits existing neurons into multiple off-springs 
to achieve progressive loss minimization, thus finding novel architectures by gradually growing the neural network.  
However, this method was not specifically tailored for designing energy-efficient networks, and is computationally expensive on large-scale benchmarks. 
In this work, we substantially improve \citet{splitting2019} in two significant ways: 
1) we 
 incorporate the energy cost of splitting different neurons 
 to better guide the splitting process, thereby discovering more energy-efficient network architectures; 
2) we substantially speed up the splitting process of   \citet{splitting2019}, which requires expensive eigen-decomposition, 
by proposing a highly scalable Rayleigh-quotient stochastic gradient algorithm.   
Our fast algorithm allows us to reduce the computational cost of splitting to the same level of typical back-propagation updates and enables efficient implementation on GPU. 
Extensive empirical results show that our method can train highly accurate and energy-efficient networks on challenging datasets such as ImageNet, improving a variety of baselines, including the pruning-based methods and expert-designed architectures. 
\end{abstract}

 \section{Introduction}
\vspace{-1em}
Deep neural networks (DNNs)
have demonstrated remarkable performance in solving
various challenge problems such as image classification \citep[e.g.][]{simonyan2014very, he2016deep, huang2017densely}, 
object detection \citep[e.g.][]{he2017mask}
and language understanding \citep[e.g.][]{devlin2018bert}. 
Although large-scale deep networks have good empirical performance, 
their large sizes 
cause slow computation and high energy cost in the inference phase.
%
This imposes a great challenge for 
improving the applicability of deep networks to more real-word domains, 
especially on mobile and edge devices where the computation and energy budgets are highly limited.
It is of urgent demand to develop practical approaches 
for automatizing the design of \emph{small, highly energy-efficient} DNN architectures 
that are still sufficiently accurate for real-world AI systems. 


Unfortunately, neural architecture optimization has been known to be notoriously difficult. 
Compared with the easier task of learning the parameters of DNNs, which has been well addressed by back-propagation \citep{rumelhart1988learning}, optimizing the network structures casts a much more  
challenging discrete optimization problem, with excessively large search spaces and high evaluation cost.  
Furthermore, 
for neural architecture optimization in energy-efficient settings, 
extra difficulties arise due to strict constraints on resource usage. 
%


Recently, 
\citet{splitting2019} 
investigated similar notations of gradient descent for learning network architectures
and framed the architecture optimization problem into a \emph{continuous optimization in an infinite-dimensional configuration space}, 
on which novel notions of \emph{steepest descent} can be derived 
for incremental update of the neural architectures. 
In practice, 
the algorithm optimizes a neural network through a cycle of \emph{paramedic updating}
and \emph{splitting} phase. In the \emph{parametric updating} phase,
the algorithm performs standard gradient descent to reach a stable local minima;
in the \emph{splitting} phase,
the algorithm expands the network 
by splitting a subset of exiting neurons into several off-springs in an optimal way.  
A key observation is that the previous local minima  
can be turned into a saddle point in the new higher-dimensional space induced by splitting that can be escaped easily; 
thus enabling implicitly architecture space exploration and achieving monotonic loss decrease.


However, the splitting algorithm in \citet{splitting2019} treats each neuron equally, without taking into account the different amount of energy consumption incurred by different neurons, 
thus finding models that may not be applicable in resource-constrained environments.
To close the gap between DNNs design via splitting and energy efficiency optimization, we propose an energy-aware splitting procedure 
that improves over  \citet{splitting2019} by 
explicitly incorporating energy-related metrics to guild the splitting process.

Another practical issue of 
\citet{splitting2019} is that it requires eigen-computation of the \emph{splitting matrices},
which causes a time complexity of $\obig(nd^3)$ and space complexity of $\obig(nd^2)$ when implemented exactly, where 
$n$ is the number of neurons in the network, and $d$ is the dimension of the weights of each neuron. 
This makes it difficult to implement the algorithm on GPUs for modern neural networks with thousands of neurons, mainly due to the explosion of GPU memory, thus prohibiting efficient parallel calculation on GPUs. 
In this work, we address this problem by 
proposing a fast gradient-based approximation of \citet{splitting2019},  
which reduces the time and space complexity to $\obig(nd^2)$ and $\obig(nd)$, respectively.
Critically, our new fast gradient-based approximation can be efficiently implemented on GPUs, hence making it possible to split very large networks, such as these for ImageNet. 


Our method achieves promising empirical results on challenging benchmarks. 
Compared with prior art pruning baselines that improve the efficiency by removing the least significant neurons \citep[e.g.][]{liu2017learning, li2016pruning, gordon2018morphnet}, 
our method produces a better accuracy-flops trade-off on CIFAR-100.
On the large-scale ImageNet dataset, 
our method finds more flops-efficient network architectures 
that achieve 1.0\% and 0.8\% improvements in 
top-1 accuracy compared with prior expert-designed MobileNet \citep{howard2017mobilenets} and MobileNetV2 \citep{sandler2018mobilenetv2}, respectively. 
The gain is even more significant on the low-flops regime.

\section{Splitting Steepest Descent}

Our work builds upon a recently proposed 
\emph{splitting steepest descent} approach \citep{splitting2019}, 
which transforms the co-optimization of 
neural architectures and parameters into a continuous optimization, solved by a generalized steepest descent on a functional space. 
To illustrate the key idea, 
assume the neural network structure is specified by a set of size parameters $\vv m= \{m_1,\ldots, m_K\},$ where each $m_k$ denotes the number of neurons in the $k$-th layer, or the number of a certain type of neurons. 
Denote by $\Theta_{\vv m}$ 
the set of possible parameters of networks of size $\vv m$, 
then $\Theta_{\vv\infty} = \cup_{\vv m\in \mathbb{N}^K} \Theta_{\vv m}$, which we call the  configuration space, is the space of all possible neural architectures and parameters.  

In this view, learning parameters of a fixed network structure  
is minimizing the loss inside  an individual sub-region  $\Theta_{\vv m}$. 
 In contrast, optimizing in the overall configuration space $\Theta_{\vv\infty}$ admits the co-optimization of both architectures and parameters. 
The key observation is that the optimization in $\Theta_{\vv\infty}$ is in fact continuous (despite being infinite dimensional), 
for which (generalized) steepest descent procedures can be introduced to yield  efficient and practical algorithms. 

\begin{figure*}
\centering
\begin{tabular}{ccc}
\setlength\tabcolsep{1pt}
\hspace{-0.5em}
\includegraphics[width=0.26\textwidth]{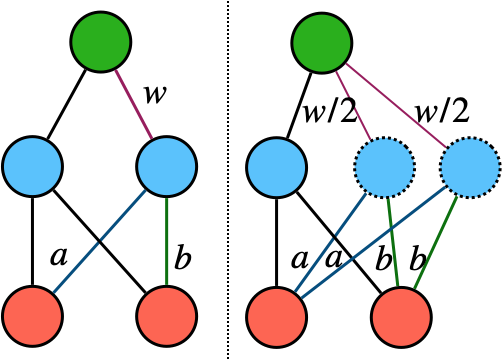} &
\hspace{-0.2em}
\includegraphics[width=0.32\textwidth]{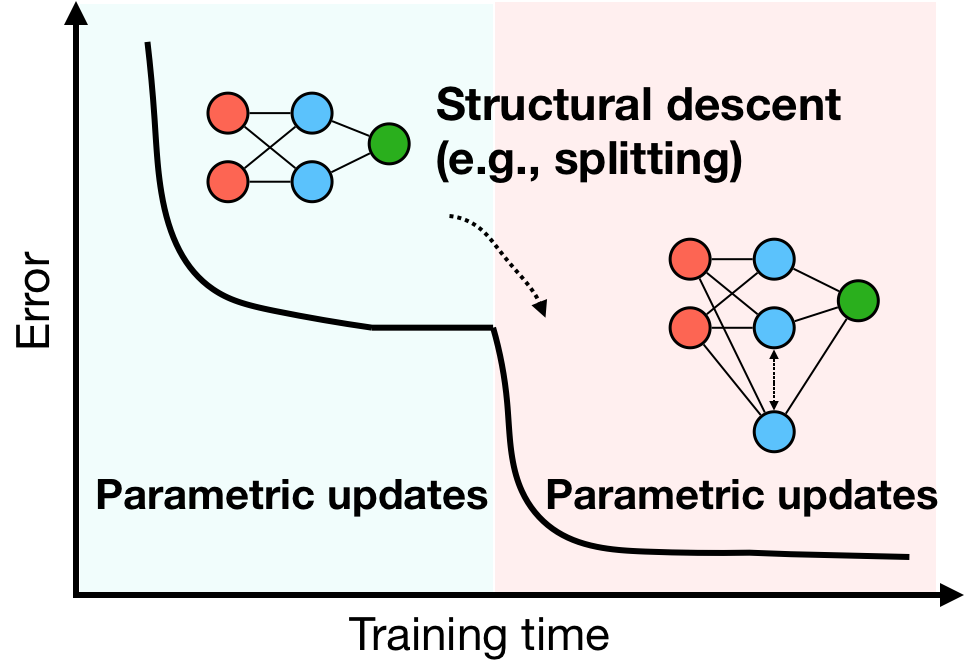} & 
\hspace{-1.2em}
\includegraphics[width=0.35\textwidth]{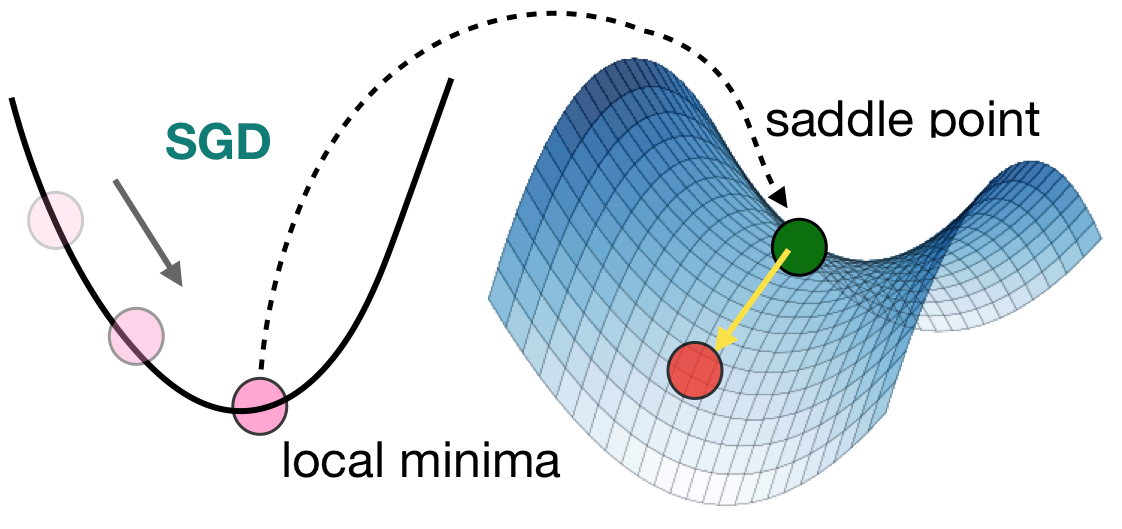} 
\\
 (a) 
 & (b) & 
 (c) 
\end{tabular}
\caption{
(a) Splitting one neuron into two off-springs.  
(b) Steepest descent on the overall architecture space 
consists of both standard gradient descent on the parameters (with fixed network structures), and updates of the network structures (via splitting). 
(c) The local optima in the low dimensional space 
is turned into a saddle point in a higher dimensional of the augmented networks, and hence can be escaped by our splitting strategy, yielding monotonic decrease of the loss. 
}
 \label{fig:intuition}
\vspace{-1em}
\end{figure*}

In particular, \citet{splitting2019} considered a \emph{splitting steepest descent} on $\Theta_{\vv\infty}$, which  
consists of two phases:  
1) the \emph{parametric descent} inside each $\Theta_{\vv m}$ with a fixed network structure $\vv m$, which 
 reduces to the typical steepest descent on parameters, 
 and 2)  the \emph{architecture descent} crossing the boundaries of different sub-regions  $\Theta_{\vv m}$,
 which, in the case of \citet{splitting2019}, corresponds to 
 ``growing'' the network structures by \emph{splitting} 
 a set of critical neurons to multiple off-springs (see Figure~\ref{fig:intuition}a). 

From the perspective of non-convex optimization, the architecture descent across boundaries of $\Theta_{\vv m}$ 
can be viewed  
as escaping saddle points in the configuration space. 
As shown in Figure~\ref{fig:intuition}b,  
when the parametric training inside a fixed $\Theta_{\vv m}$ gets saturated, 
architecture descent allows us to escape local optima by jumping into a higher dimensional sub-region of a larger network structure. 
The idea is that the local optima inside $\Theta_{\vv m}$ can be turned into a saddle point when viewed from the higher dimensional space of larger networks (Figure~\ref{fig:intuition}c), 
which is escaped using splitting descent. 

\renewcommand{\gM}{\Phi}

\paragraph{Escaping local minima via splitting}
\label{sec:single}
It requires to fix a proper notion of distance on $\Theta_{\vv\infty}$ in order to 
derive a steepest descent algorithm. 
In \citet{splitting2019}, 
steepest descent with $\infty$-Wasserstein distance was considered, 
which is shown to naturally correspond to 
the practical procedure of splitting neurons. 
Here 
we only introduce the intuitive idea 
from the practical perspective of optimally splitting neurons.   The readers are referred to \citet{splitting2019} for more theoretical discussion.

Consider the simplest case of splitting a single neuron. 
Let $\sigma(\theta, x)$ be a neuron inside a neural network that we want to learn from data, where $\theta$ is the parameter of the neuron and $x$  its input variable. 
Assume the loss function of $\theta$ has a general form of 
\begin{align}\label{equ:Ltheta}
L(\theta) :=  \E_{x\sim\mathcal D}[\Phi(\sigma(\theta, x))], 
\end{align}
where $\mathcal D$ is the data distribution, and $\Phi(\cdot)$ is the map from the output of the neuron to the final loss. %
{In this work, the word ``neuron'' broadly refers to 
repeatable modules in neural networks, such as 
the typical hidden neurons, filters in CNNs.}

Assume we have achieved a stable local optimum of $L(\theta)$, that is, $\nabla L(\theta) = 0$ and $\nabla_{\theta\theta} L(\theta) \succ 0 $, 
 so that we can not further decrease the loss by local descent on the parameters. 
%
In this case, splitting steepest descent enables further descent by introducing more neurons via splitting. 
Specifically, we split $\theta$ into $m$ off-springs  $\vv\theta := \{\theta_i\}_{i=1}^m$, and replace the neuron $\sigma(\theta ,x)$ with a 
weighted sum of the off-spring neurons $\sum_{i=1}^m w_i \sigma(\theta_i, x)$, where $\vv w:= \{w_i\}_{i=1}^m$  
 is a set of positive weights assigned on each of the off-springs, and satisfies $\sum_{i=1}^m w_i = 1$, $w_i > 0$.  
   See Figure~\ref{fig:intuition}a for an illustration. 
This yields an augmented loss function on $\vv\theta$ and $\vv w$: 
\begin{align} \label{equ:LLtheta}
\Lm(\vv\theta, \vv w) :=  
\E_{x\sim \mathcal D}\left [\Phi \left (  \sum_{i=1}^m  w_i\sigma(\theta_i, x)   \right ) \right]. 
\end{align}
It is easy to see that if we set 
 $\theta_i = \theta$ for all the off-springs, the network remains unchanged. 
Therefore, as we change $\theta_i$ in a small neighborhood of $\theta$,  it introduces a smooth change on the loss function. 
Splitting steepest descent is derived 
by considering the optimal splitting strategies to achieve the steepest descent on loss in a small neighborhood of the original parameters. %


\paragraph{Deriving splitting steepest descent}
Derive the optimal splitting strategy involves deciding the number of off-springs $m$, the values of the weights $\{w_i\}$
and the parameters for the off-springs $\{\theta_i\}$.
In \citet{splitting2019},
this is formulated into the following optimization problem: 
\begin{align}
\!\!\!\!\min_{m, \vv\theta, \vv w} \bigg\{ \Lm(\vv\theta, \vv w) - L(\theta) ~~~
\text{s.t.}~~~ ||\theta_i - \theta|| \le \epsilon, ~ \sum_{i=1}^m w_i = 1, ~~w_i > 0, ~~\forall~i\in [m], ~~m \in \mathbb{N}_+  \bigg\}, 
\label{equ:splitting_loss}
\end{align}
where the  parameters $\vv\theta:=\{\theta_i\}_{i=1}^m$ of the off-springs are restricted 
within an infinitesimal $\epsilon$-ball of the original parameter $\theta$, 
that is, $ ||\theta_i - \theta|| \le \epsilon$, 
with $\epsilon$ a small positive step size parameter.
Note that the number of off-springs $m$ is also optimized, yielding an infinite  dimensional optimization. 

Fortunately, when $\epsilon$ is very small, 
the optimum of \Eqref{equ:splitting_loss} is 
achieved by either $m = 1$ (no splitting) or $m = 2$ (two off-springs). 
The property of the optimal solution is characterized (asymptotically) by 
the following key \emph{splitting matrix} $S(\theta)$, 
\begin{align*}
    S(\theta) = \E_{x\sim \gD} \bigg[\nabla_\sigma\gM(\sigma(\theta, x))  \nabla_{\theta\theta}^2 \sigma(\theta, x)\bigg], 
\end{align*}
which is a symmetric $d\times d$ matrix ($d$ is the dimension of $\theta$). 
The optimum of \Eqref{equ:splitting_loss}, when $L(\theta)$ reaches a stable local optimum (i.e., $\nabla_\theta L(\theta) =0$, $\nabla_{\theta\theta}L(\theta) \succ 0$), is determined by $S(\theta)$ via 
\begin{align}
&\min_{m, \vv\theta, \vv w} \bigg\{ \Lm(\vv\theta, \vv w) - L(\theta) \bigg\}
= \frac{\epsilon^2}{2}\min\left \{\lambda_{\min}(S(\theta)), ~ 0\right\} ~+~ \obig(\epsilon^3), 
\label{eq:splitting_vanilla}
\end{align}
 where $\lambda_{\min}(S(\theta))$ denotes the  minimum eigenvalue of $S(\theta),$ 
 and it is called the  \emph{splitting index}. 

When $\lambda_{\min}(S(\theta)) > 0$, 
the loss can not be improved by any splitting strategies following \eqref{eq:splitting_vanilla}.  
When $\lambda_{\min} (S(\theta)) < 0$, 
the maximum decrease of loss, which equals $\epsilon^2\lambda_{\min}(S(\theta))/2$, 
can be achieved by a simple strategy of \emph{splitting the neuron into two copies with equal weights}, whose parameters are updated along the minimum eigen-vectors $v_{\min}(S(\theta))$ of $S(\theta)$, that is, 
\begin{align} 
m = 2, &&
\theta_1 = \theta + \epsilon v_{\min}(S(\theta)), && 
\theta_2 = \theta - \epsilon v_{\min}(S(\theta)), &&
w_1 = w_2 = 1/2.
\label{eq:binasplitting}
\end{align}
In this case, splitting allows us to escape the parametric local optima to enable further improvement. 

\paragraph{Splitting deep neural networks}
As shown in \citet{splitting2019}, 
the result above can be naturally extended to 
more general cases when we need to split multiple neurons in deep neural networks.  
Consider a 
neural network with $n$ neurons $\theta^{[1:n]} = \{\theta^{1}, \cdots, \theta^{n}\}$. 
Assume we split a subset $A$ of neurons 
with the optimal strategy in \Eqref{eq:binasplitting} following their own splitting matrices, the improvement of the overall loss equals the sum of individual gains 
$
G(A) = \sum_{\ell \in A} \lambda_{\min}(\ell),
$
where $\lambda_{\min}(\ell):= \lambda_{\min}(S_\ell(\theta^\ell))$ 
denotes the minimum eigenvalue of the splitting matrix $S_\ell(\theta^\ell)$ associated with neuron $\ell$. 
Therefore, 
given a budget of splitting at most a given number of neurons, 
the optimal subset of neurons to split are the top ranked neurons with the smallest, and negative minimum eigenvalues. 
Overall, the splitting descent in \citet{splitting2019} alternates between parametric updates with fixed network architectures, and splitting top ranked neurons to augment the  architectures, until a stopping criterion is reached. 

\section{Neural architecture optimization via energy-aware splitting}
\label{sec:energy-splitting}

The method above allows us to select the best subset of neurons to split to yield the steepest descent on the loss function. 
In practice, however, 
splitting different neurons incurs a different amount of increase on the model size,  computational cost, and physical energy consumption.
%
For example, 
splitting a neuron connecting to a large number of inputs and outputs increases the size and computational cost of the network much more significantly than splitting the neurons with fewer inputs and outputs. 
{In practice, convolutional layers close to inputs often have larger feature maps which lead to a high energy cost, and layers 
closer to outputs have smaller feature maps and hence lower computational cost. }
A better splitting strategy should  take the cost of  different neurons into account. 

To address this problem, we propose to explicitly incorporate the  
energy cost to better guide the splitting process.
Specifically, for a neural network with $n$ neurons, we 
propose to decide the optimal splitting set by solving the following constrained optimization: 
\begin{align}
\min_{\vv \beta} \sum_{\ell=1}^n \beta_\ell \lambda_{\min}(\ell), ~~~~\text{s.t.}~~\sum_{\ell=1}^n e_\ell \beta_\ell \le e_{budget}, 
~~\beta_\ell \in \{0, 1\}, 
~~\forall \ell  
\label{eq:energy_aware_splitting}  
\end{align}
Here $\vv \beta\in \R^n$ is a binary mask, with $\beta_\ell$
indicates whether the $\ell$-th neuron should be split ($\beta_\ell=1$) or not ($\beta_\ell=0$),
and $e_\ell$ represents the cost of splitting at the current iteration.
We search for the optimal subset of neurons that yields the largest descent on the loss (in terms of the splitting index), while incurring a total energy cost no larger than a budget threshold $e_{budget}$. 
This optimization \Eqref{eq:energy_aware_splitting} is a standard 
\emph{knapsack problem}. 
The exact solution of knapsack problems can be very expensive due to their NP-hardness.
In practice, we use linear programming relaxation for fast approximation by relaxing the integrality constrains to linear constrains such that $\beta_\ell \in [0, 1], \forall \ell$.
The continuous relaxation could then be solved using standard linear programming tools efficiently \citep{dantzig1998linear}. Finally, we define the optimal splitting set 
$A := \{\beta_\ell > 0.9, \forall \ell\} $. For each neuron in $A$, we split it into two equally 
weighted off-springs along their splitting gradients, following~\Eqref{eq:binasplitting}.


In this work, we take $e_\ell$ to be the energy cost, 
and estimate it by the increase of flops if we split the $\ell$-th neuron starting from the current network structure.  
%
Note that the cost of splitting the same neuron changes when the network size changes across iterations.  
Therefore, we re-evaluate the cost of every neuron at each splitting stage, 
based on the architecture of the current network. 





\section{
Fast Splitting with Rayleigh-Quotient Gradient Descent }
\label{sec:fast_grad_approx}

A practical issue of splitting steepest descent  
is the high computational cost of 
the eigen-computation of the splitting matrices. 
The time complexity of evaluating all splitting indexes is  $\obig(nd^3)$. 
Here $n$ is the number of neurons and $d$ is the dimension of each neuron. 
Meanwhile, the space complexity is $\obig(nd^2)$.
Although this is manageable for networks with small or moderate sizes,
an immediate difficulty for modern deep networks with thousands of high-dimensional neurons ($\sim\!1000$) is that we are not able to store all splitting-matrices on GPUs,
which necessities slow calculation on CPUs.
It is desirable to further speed up the calculation for very large scale problems.
%
In this section, 
we propose  
an approach for computing the splitting indexes and gradients \emph{without explicitly expanding the splitting matrices}, based on fast (stochastic) gradient descent on the Rayleigh quotient. 

\paragraph{Rayleigh-Quotient Gradient Descent}  
The key idea is to note that the minimum eigenvalues and eigenvectors of a matrix $S\in \R^{d\times d}$ can be obtained by minimizing Rayleigh quotient \citep{parlett1998symmetric}, 
\begin{align} 
\lambda_{min} 
= \min_{v}\left \{ \mathcal R_S(v) := \frac{ v^\top S v }{v^\top v} \right\}, 
&&
v_{min} \propto \argmin_{v}  \mathcal R_S(v), 
\label{eq:rayley_ratio}
\end{align}
 which can be solved using gradient descent or other numerical methods.  
 Although this problem is non-convex, $v_{min}$ can be shown to be the unique global minimum of $R(v)$, 
 and all the other stationary points, 
 corresponding to the other eigenvectors, 
 are saddle points and can be escaped with random perturbation. 
 Therefore, stochastic or noisy gradient descent on $R(v)$ is expected to converge to $v_{min}$. 
 The gradient of $R(v)$ w.r.t. $v$ can be written as follows, 
 $$
 \nabla_v \mathcal R_S(v)
 = 2\norm{v}^{-2} \left ( S v - \mathcal R_S(v) v \right)
 \propto S v - \mathcal R_S(v) v, 
 $$
 which depends on $S$ only through the matrix-vector product $S v$. 
 A significant saving in computation can be obtained by directly calculating $Sv$ at each iteration, without explicitly expanding the whole matrix. 
 This can be achieved by the following auto-differentiation trick. 
 
 \paragraph{Automatic Differentiation Trick} 
 Recall that the splitting matrix of a single neuron is 
 $S(\theta) = \E_{x\sim \gD}[\nabla_\sigma \Phi(\sigma(\theta, x))\nabla^2_{\theta\theta}\sigma(\theta,x)]$.  
To calculate $S(\theta) v$ for any vector $v\in \RR^d$, we construct the following auxiliary function, 
\begin{align*} 
F(\eta)   = \E_{x\sim \gD}[\gM(\sigma(\theta, x) +  \textcolor{blue}{\eta^\top \nabla_{\theta\theta}^2 \sigma(\theta,x) v} ) ) ],~~~~~~ \eta \in \RR^d,  
\end{align*}
with which it is easy to show that $S(\theta)v = \nabla_\eta F(0)$.  
Here $F(\eta)$ corresponds to 
simply adding an extra term on the top of the neuron's output and can be constructed conveniently.
\begin{wrapfigure}{r}{0.3\textwidth}
\begin{small}
\centering
\begin{tabular}{c}
\includegraphics[height =0.25\textwidth]{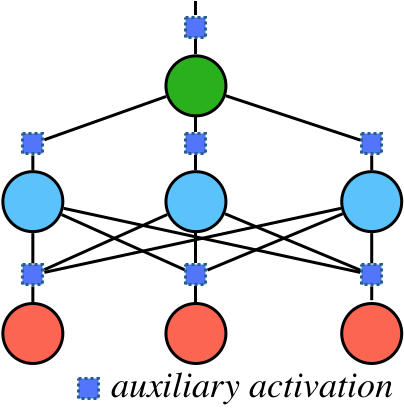} 
\vspace{-1em}
\end{tabular}
\caption{Illustration of the auto-differentiation trick with auxiliary activation.}
\label{fig:aux_gradient}
\end{small}
\vspace{-3em}
\end{wrapfigure}

In the case of deep neural networks with $n$ neurons $\{\theta^{\ell}\}_{\ell=1}^n$, 
we can calculate all the matrix-vector product $\{g_\ell := S_\ell(\theta^\ell) v_\ell\}_{\ell=1}^n$ for all the neurons  jointly with a single differentiation process. 
More precisely, for each neuron $\theta^\ell$, we can add a term $\textcolor{blue}{\eta_\ell^\top \nabla_{\theta\theta}^2 \sigma_\ell(\theta^{\ell},x) v_\ell}$ (denoted as \emph{auxiliary activation}) on its own output (see Figure~\ref{fig:aux_gradient}). Thus, we obtain a joint function $F(\eta_1, \ldots, \eta_n)$, for which it is easy to see that $\nabla_{\eta_\ell} F(\eta_1, \ldots, \eta_n) = g_\ell$, $\forall \ell$.  
Therefore, simply differentiating $F(\eta_1, \ldots, \eta_n)$ allows us to obtain all $\{g_\ell\}$ simultaneously. 

\paragraph{Stochastic Gradient on Rayleigh quotient} 
Note that we still need to average over the whole dataset $\gD$ to measure the Rayleigh quotient gradients $\{g_\ell\}$, this is computationally expensive in the case of big data. 
However, we can conveniently address this by approximating $\{g_\ell\}$
with subsampled mini-batches $\gB$. In the case of single-neuron networks, that is, 
$$
Sv = \nabla_\eta \hat{F}(\eta), ~~\text{with}~~\hat{F}(\eta) = \frac{1}{|\gB|} \sum_{i=1}^{|\gB|}\bigg[ \Phi\bigg( \sigma(\theta, x_i) + \eta^\top\nabla_{\theta\theta}^2 \sigma(\theta, x_i) v \bigg)\bigg].
$$
Assume we sweep the training data $T$ times to train the Rayleigh-Quotient to convergence (see \Eqref{eq:rayley_ratio}). 
In this way, the splitting time complexity for 
approximating all splitting indexes and gradients would be only $\gO(Tnd^2)$ ($T$ is often a small constant).
%
More importantly, a significant advantage of our gradient-based approximation is that the space complexity is only $O(nd)$. In this way, all calculation could be efficient performed on GPUs. This given us an algorithm for splitting that is almost as efficient as back-propagation.





\begin{algorithm*}[t] 
\caption{Energy-aware neural architecture optimization with fast splitting steepest descent} 
\begin{algorithmic} 
\STATE Starting from a small base network (viewed as the ``seed''), we gradually grow the neural network by alternating between the following two phases until an energy constrain reached:  
\STATE \textbf{1. Parametric Updates}:
    Optimize neuron weights using standard methods (e.g., SGD) until no further improvement can be made by only updating parameters. 
    \vspace{.3\baselineskip} 
    
    \STATE \textbf{2. Splitting Neurons}:  
    \begin{enumerate}
    \item[(a)] Computing the splitting index of each neuron using gradient-based approximation (Section~\ref{sec:fast_grad_approx});
     \item[(b)] Finding the optimal set of neurons to split by solving the energy-aware allocation problem in ~\Eqref{eq:energy_aware_splitting}; 
    \item[(c)] For each neuron selected above (step 2a), split it into two equally weighted off-springs along their splitting gradients, following~\Eqref{eq:binasplitting}. 
    \end{enumerate}
\end{algorithmic}
\label{alg:main}  
\end{algorithm*}

\paragraph{Overall Algorithm}
Our overall algorithm in shown in Algorithm~\ref{alg:main},
which improves over \citet{splitting2019} 
by offering much lower time and space complexity, 
and the flexibility of incorporating energy and other costs of different neurons. 
It can be implemented easily using modern deep learning frameworks such as Pytorch \citep{paszke2017automatic}. 
Our code is available at 
{\url{https://github.com/dilinwang820/fast-energy-aware-splitting}}.

\section{Experiments}
We apply our method to split small variants of MobileNetV1 \citep{howard2017mobilenets} and MobileNetV2 \citep{sandler2018mobilenetv2}, 
on both CIFAR-100 and ImageNet dataset. 
We show our method finds networks that are more accurate and also more energy-efficient 
compared to expert-designed architectures 
and pruned models. 

\paragraph{Settings of Our Algorithm} 
In  all our tests of our Algorithm~\ref{alg:main}, 
we restrict the increase of the energy cost to
be smaller than a budget $e_{budget}$ at each splitting stage.
We set $e_{budget}$ adaptively to be proportional to the total flops 
of the current network such that 
the flops of the augmented network obtained by splitting cannot exceed 
$1+\alpha$ times of the previous one.
We denote by $\alpha$ the growth ratio and 
set $\alpha=0.5$ unless otherwise specified.

For our fast \emph{splitting indexes} approximation (see section~\ref{sec:fast_grad_approx}),
we set batch size $|\gB|$ to be 64 and use 
RMSprop \citep{tieleman2012lecture} optimizer with 0.001 learning rate.  
We find the Rayleigh-Quotient converges fast in general: 
for small CIFAR-10/100 datasets, we train 10 epochs (T=10); 
for the large-scale ImageNet set, we find a small T (=2) is sufficient. 



\subsection{
Testing Importance of Energy-Aware Splitting (Results on CIFAR-10)}
To study the importance of our energy-aware splitting, 
we compare our method (denoted as \emph{splitting (energy-aware)}) to \citet{splitting2019} (denoted as \emph{splitting (vanilla)}), which doesn't use energy metrics to guide the splitting process. In this experiment, 
we apply both splitting algorithms to grow a variant of small version of MobileNets \citep{howard2017mobilenets} trained on the CIFAR-10 dataset, in order to test the importance of using energy cost for splitting. 


\paragraph{Settings} 
We test our algorithm on two variants of MobileNet, 
each of which 
consists one regular $3\times 3$ convolution layer, followed by $k=3$ and $k=6$ MobileNet blocks \citep{howard2017mobilenets}, respectively. 
In both variants, the resolutions are reduced 3 times evenly and {one extra MobileNet block attached with a fully connected layer for classification.}  
Note that each MobileNet block consists a depthwise convolutional layer and
a pointwise convolutional layer. In our implementation, 
we only split the convolutional filters in the pointwise convolutional layers and duplicate the corresponding depthwise convolution filters accordingly during splitting.  
We start with small networks that have the 
same number of channels (=8) across all layers
to better study the behavior of how neurons are split.
We set batch size to be 256 and learning rate 0.1 for 160 epochs, with learning rate dropped 10x at 80 and 120 epochs {for the two variants ($k=3,6$), respectively}. 

\begin{figure*}[ht]
\begin{center}
\renewcommand{\tabcolsep}{2pt}
\begin{tabular}{cccc}
{\scriptsize (a) k = 3} 
& {\scriptsize (b) k = 3} 
&{\scriptsize (c) k = 6}
& {\scriptsize (d) k = 6} ~~~\\
\raisebox{0.5em}{\rotatebox{90}{{\scriptsize Test Accuracy}}}~~
\includegraphics[height =0.18\textwidth]{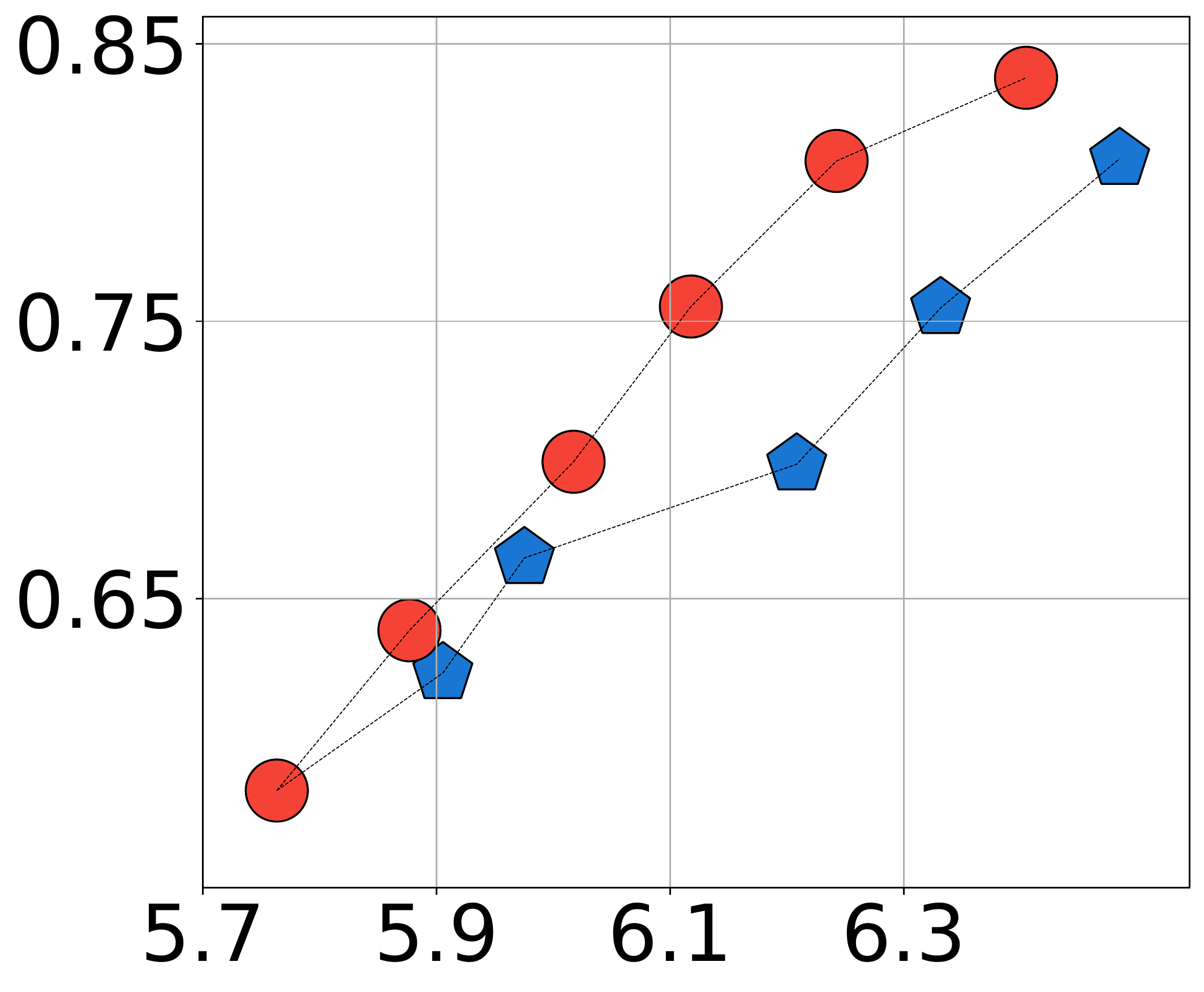} & 
\includegraphics[height =0.18\textwidth]{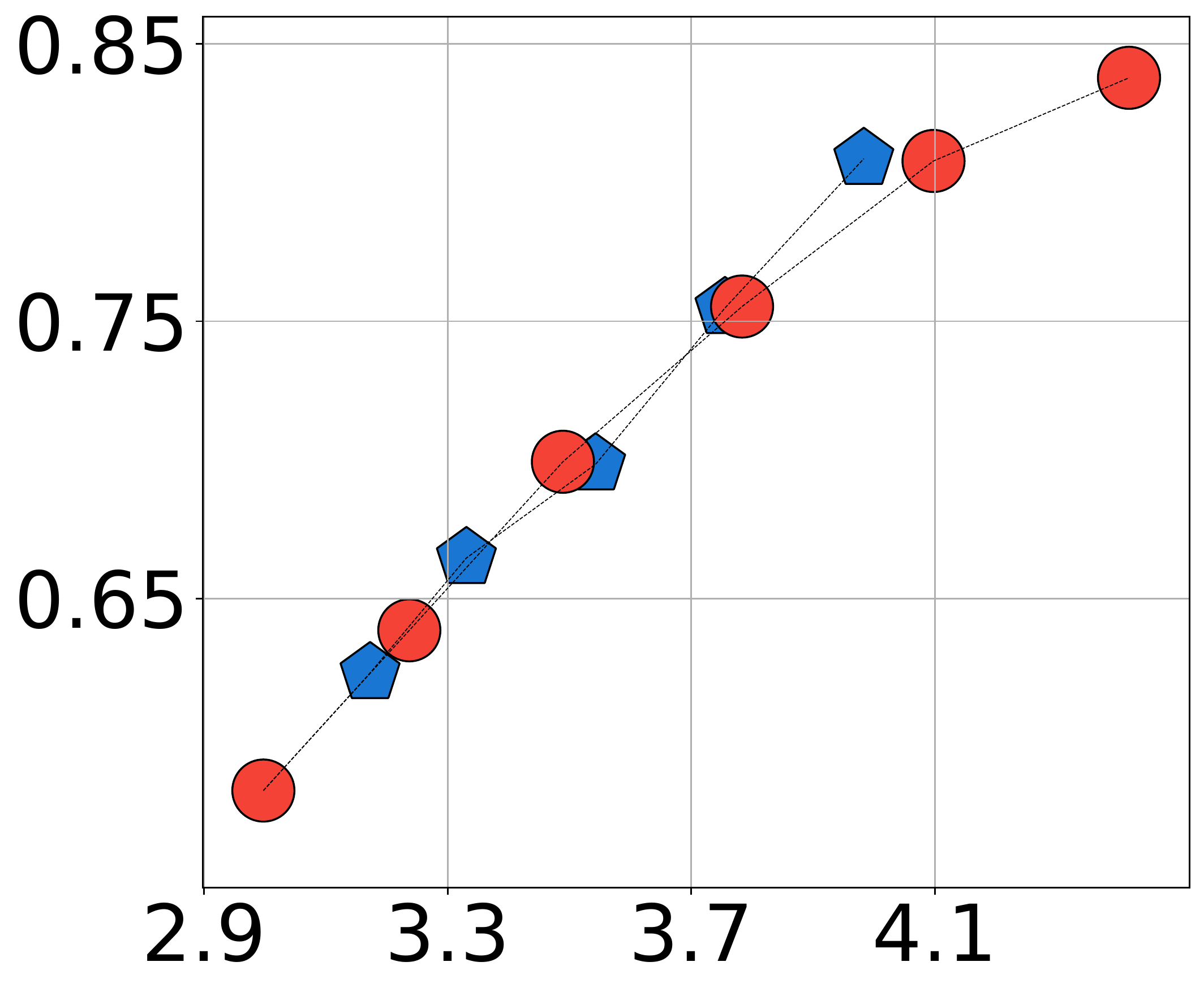} & 
\includegraphics[height =0.18\textwidth]{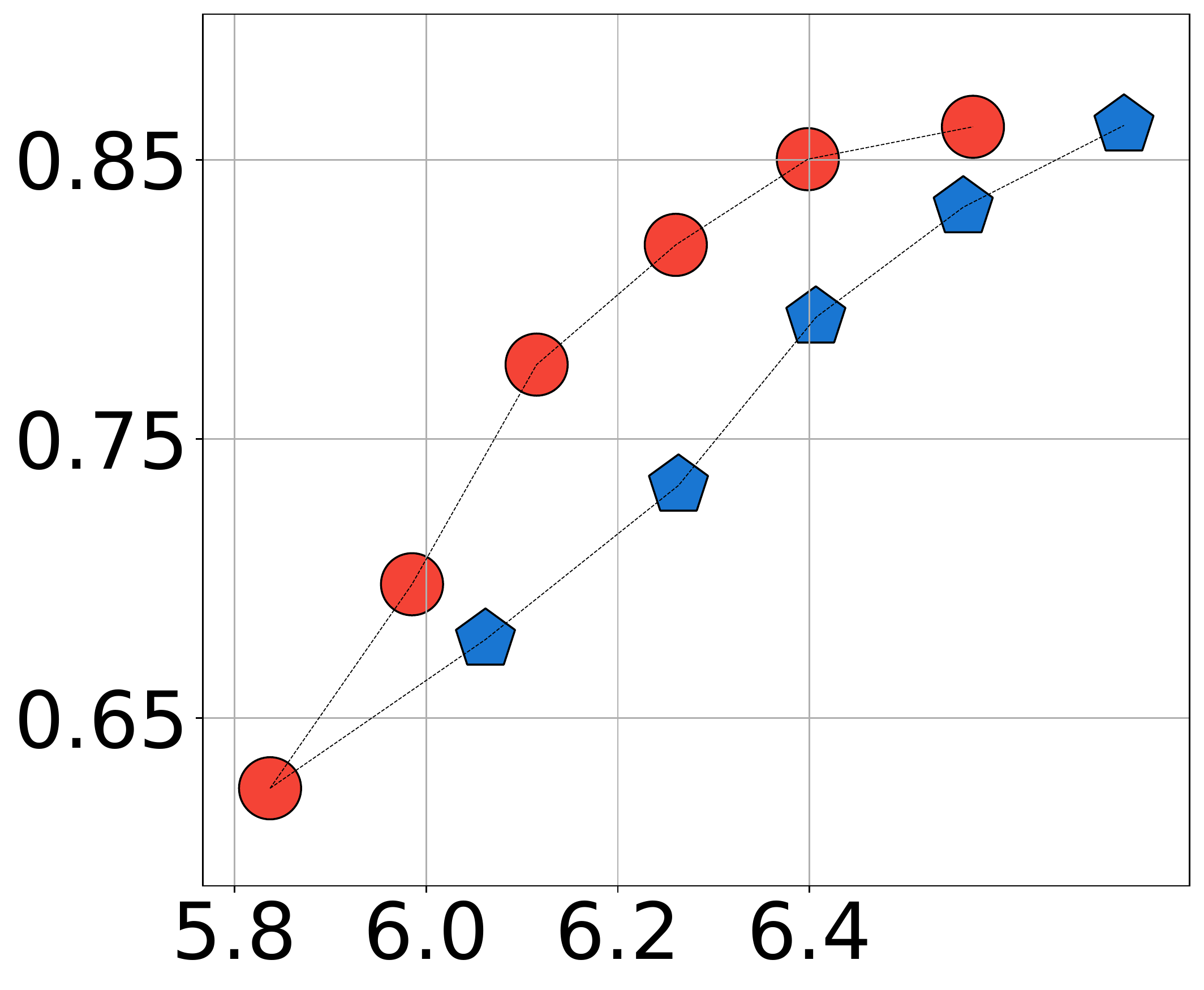} & 
\includegraphics[height =0.18\textwidth]{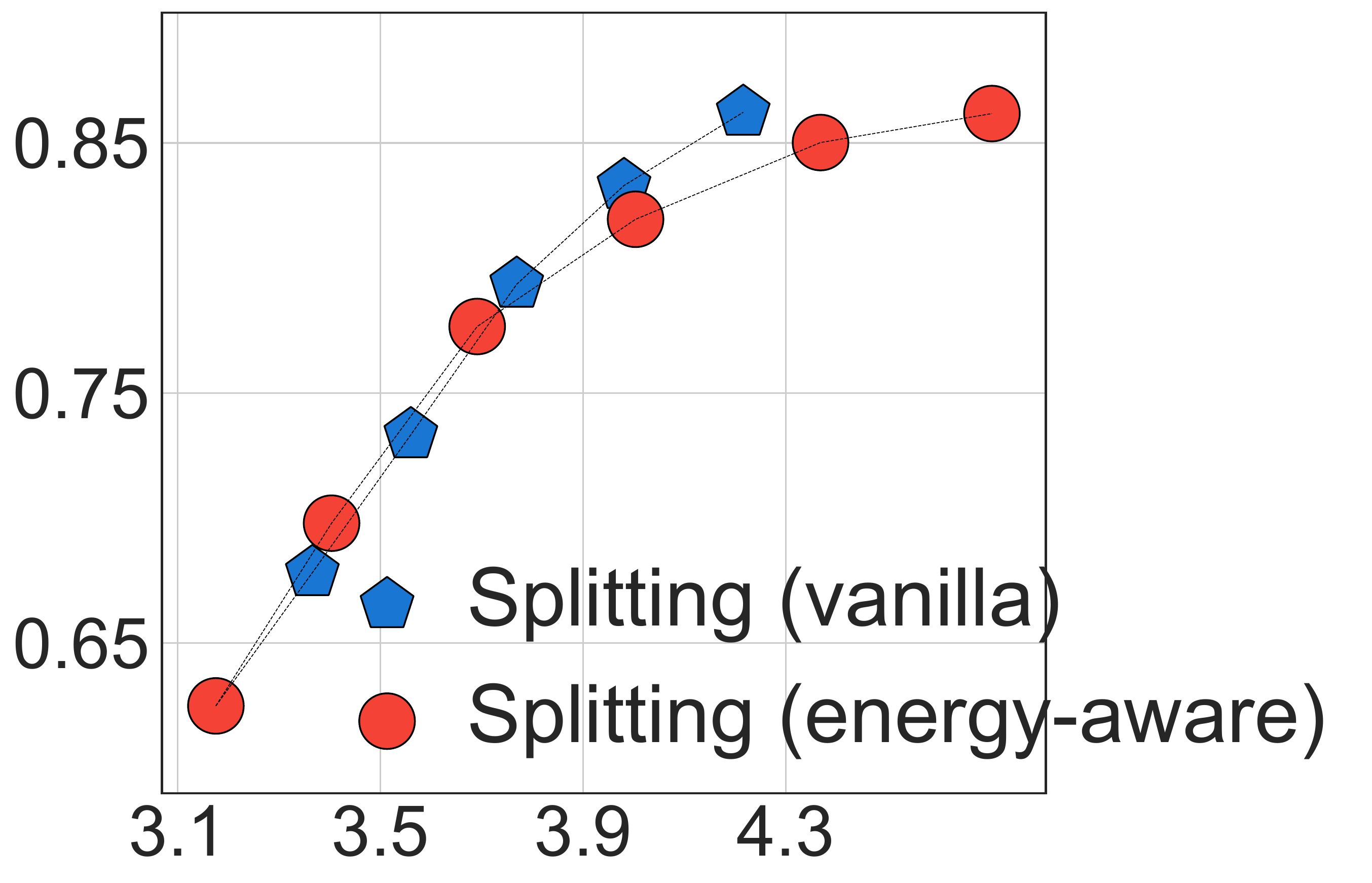} \\
{\scriptsize Log10(flops)} 
& {\scriptsize Log10(\#parameters) }
& {\scriptsize Log10(flops) }
& {\scriptsize Log10(\#parameters)}  \\
\end{tabular}
\end{center}
\caption{\small 
Comparisons between our energy-aware splitting and standard splitting in \citet{splitting2019} on CIFAR-10. 
{Results are shown for two variants of MobileNet, one with $k=3$ (4 MobileNet blocks, 9 layers in total), another with $k=6$ (7 MobileNet blocks, 15 layers in total).}
}
\label{fig:mbv1_small_cifar10}
\end{figure*}

\paragraph{Results} 
Our results are shown  in Figure~\ref{fig:mbv1_small_cifar10}, which shows that our \emph{splitting (energy-aware)} approach yields better  trade-offs of accuracy and flops 
than \emph{splitting (vanilla)} in both cases ($k=3$ and $k=6$). 
We find that  \emph{splitting (vanilla)}  does discover networks with small model size (fewer parameters, see Figure~\ref{fig:mbv1_small_cifar10} {(b)} and (d)), 
but does not yield lower energy consumption in practice. 
These results highlight the importance of using
real energy cost for guiding the splitting process in 
order to optimize for the best energy-efficiency.

\subsection{Results on CIFAR-100}
\label{sec:cifar100}
We compare our method with several {state-of-the-art} pruning baselines on the CIFAR-100 dataset. 
We also show our fast gradient-based splitting approximation in section~\ref{sec:fast_grad_approx} achieves the same accuracy as
the exactly eigen-computation, 
while significantly reducing the overall splitting time.

\paragraph{Settings} 
{We again apply splitting on a small version of MobileNet \citep{howard2017mobilenets} (with the same network topology)
to obtain a sequence of increasingly large models.} 
Specifically,  
we set the number of channels of the base model to be 
$2,4,8,8,16,16,24,24,24,24,24,24,32,32$ for each layer,
respectively.
We compare {our method} with a simple but competitive \emph{width multiplier} \citep{howard2017mobilenets} baseline, 
which prunes filters uniformly across layers (denoted as \emph{Width multiplier}) from the 
original full size MobileNet. 
We also experiment with  
three state-of-the-art structured pruning methods:
\emph{Pruning (Bn)} \citep{liu2017learning}, \emph{Pruning (L1)}  \citep{li2016pruning} and \emph{MorphNet} \citep{gordon2018morphnet}.
%
The implementation of all the baselines are based on \citet{liu2018rethinking}.
For all methods, 
we normalize the inputs using channel means and standard deviations. 
We use stochastic gradient descent with momentum 0.9, weight decay 1e-4, 
batch size 128. We set 0.1 initial learning rate for 160 epochs,
with learning rate decreased by 10x at epochs 80, 120, respectively. 
For all pruned models, we report the finetune performance 
with the same training settings.
For Morphnet, we grid search the best sparsity hyper-parameter $\lambda$
in the range [1e-8, 5e-8, 1e-9, 5e-9, 1e-10] and report the best models found.

\begin{figure*}[t]
\begin{center}
\begin{tabular}{ccc}
\raisebox{2.0em}{\rotatebox{90}{{\scriptsize Test Accuracy}}}~~
\includegraphics[height =0.24\textwidth]{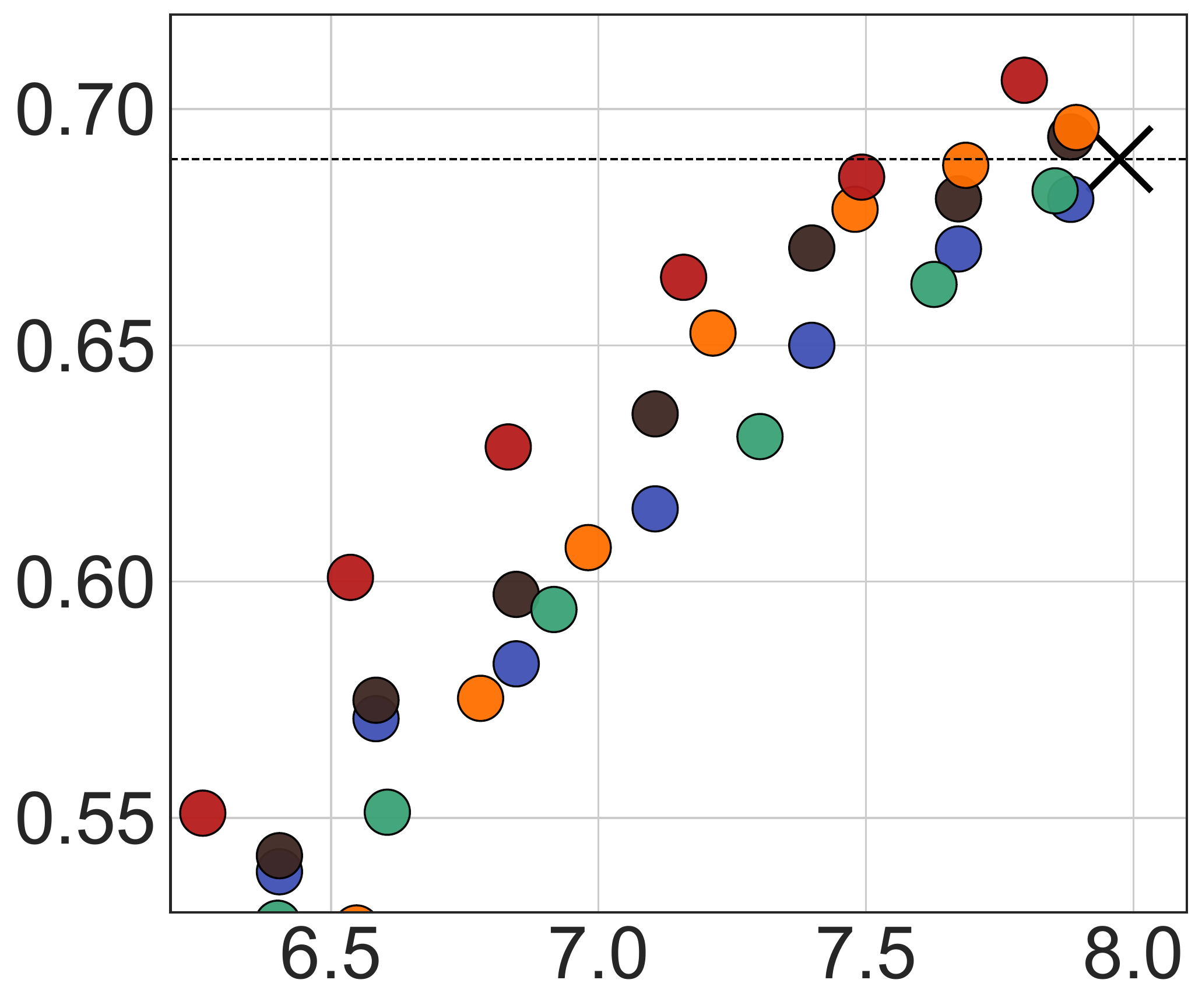} &
\hspace{-1.5em}
\raisebox{1em}{\includegraphics[height =0.14\textwidth]{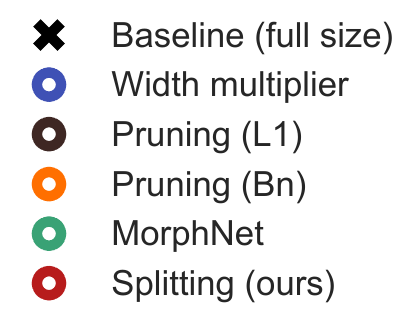}} & 
\raisebox{2.0em}{\rotatebox{90}{{\scriptsize Test Accuracy}}}~~
\includegraphics[height =0.24\textwidth]{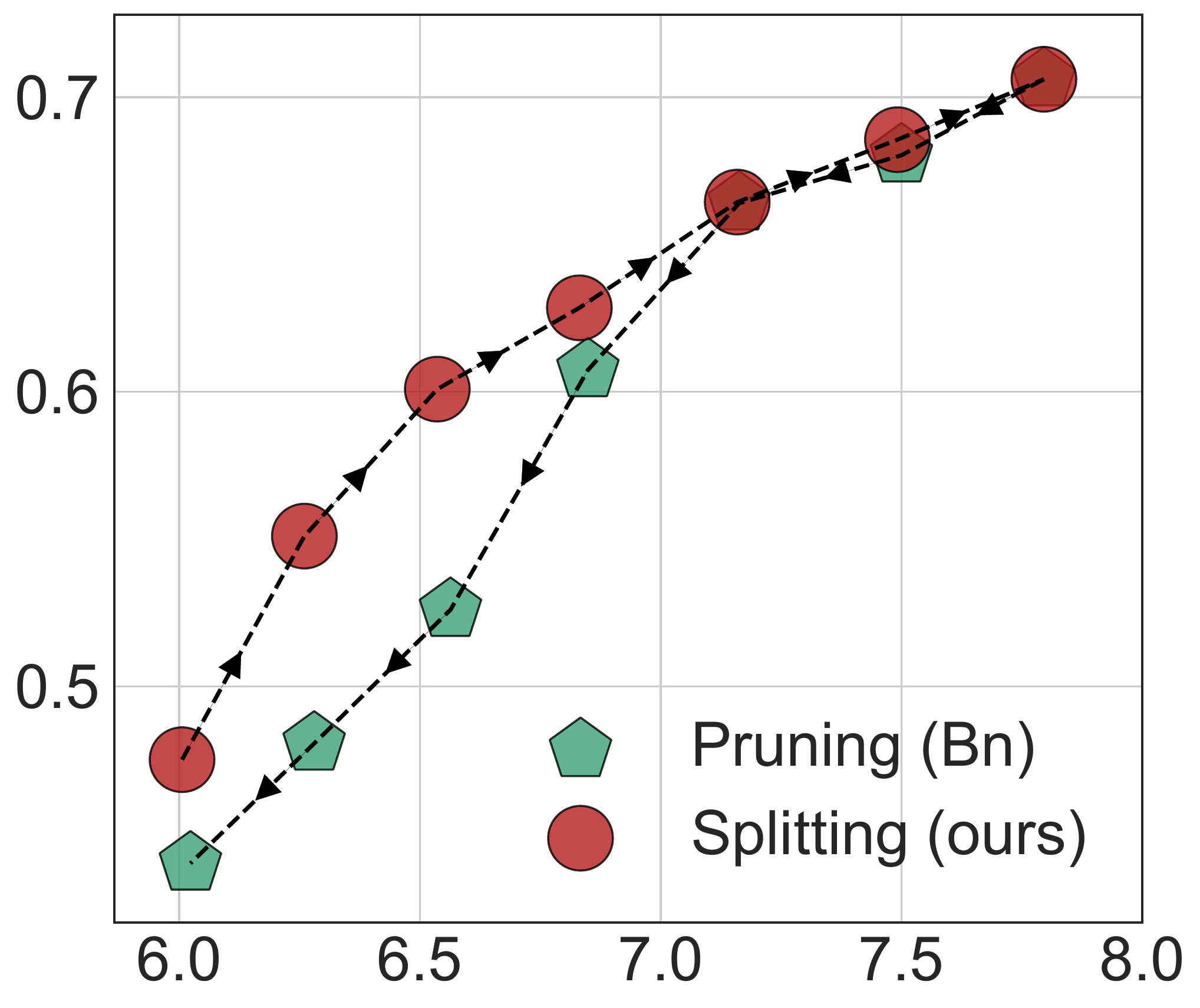} \\
{\scriptsize (a) Log10(flops)} & &
{\scriptsize (b) Log10(flops)} \\
\end{tabular}  
\end{center}
\caption{\small (a) Results on CIFAR-100 using MobileNet\citep{howard2017mobilenets};
(b) we show our energy-aware splitting approach can learn more accurate and energy-efficient (with small flops) networks than pruning methods \citep{liu2017learning}.} 
\label{fig:mbv1_cifar100}
\end{figure*}

\paragraph{Results} 
Figure~\ref{fig:mbv1_cifar100} (a) shows the results on  CIFAR-100, 
in which our method achieves the best accuracy  when targeting similar flops.
To draw further comparison between the splitting and pruning approaches, 
we prune the final network learned by our splitting algorithm  to obtain a sequence of increasingly smaller models using \emph{Pruning (Bn)} \citep{liu2017learning}. 
As shown in Figure~\ref{fig:mbv1_cifar100} (b), 
it is clear that our splitting checkpoints (red circles) 
form a better flops-accuracy trade-off curve than models obtained by pruned from the same model (green Pentagons). 
This confirms the advantage of our method in neural architecture optimization, 
especially on the low-flops regime.


\begin{wrapfigure}{r}{0.57\textwidth}
\begin{small}
\centering
\vspace{-1.5em}
\begin{tabular}{cc}
\raisebox{1.1em}{\rotatebox{90}{\scriptsize { Test Accuracy}}}~~
\includegraphics[height =0.19\textwidth]{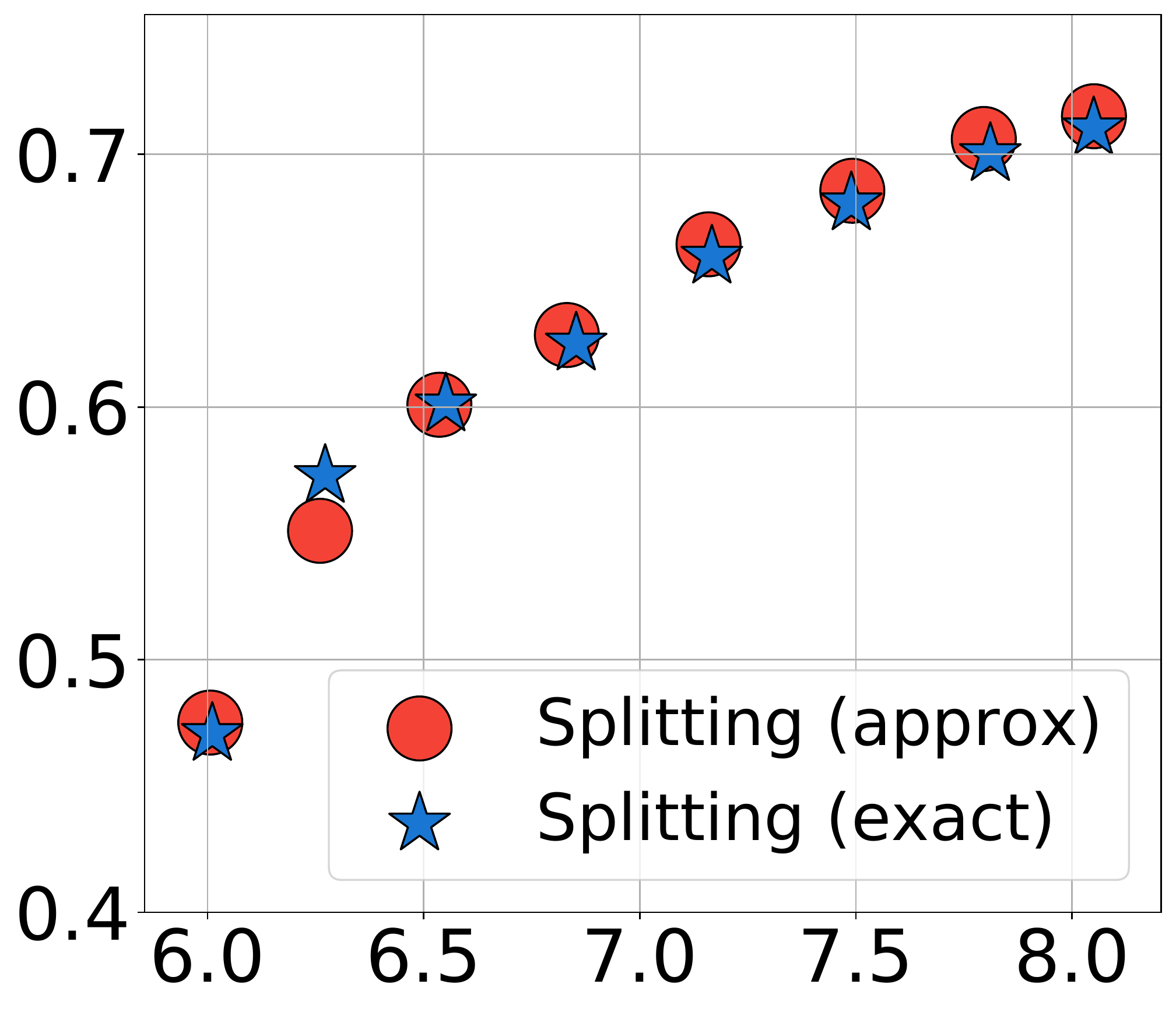} &
\hspace{-0.5em}
\raisebox{1.1em}{\rotatebox{90}{\scriptsize{ Splitting time (s)}}}~
\includegraphics[height =0.19\textwidth]{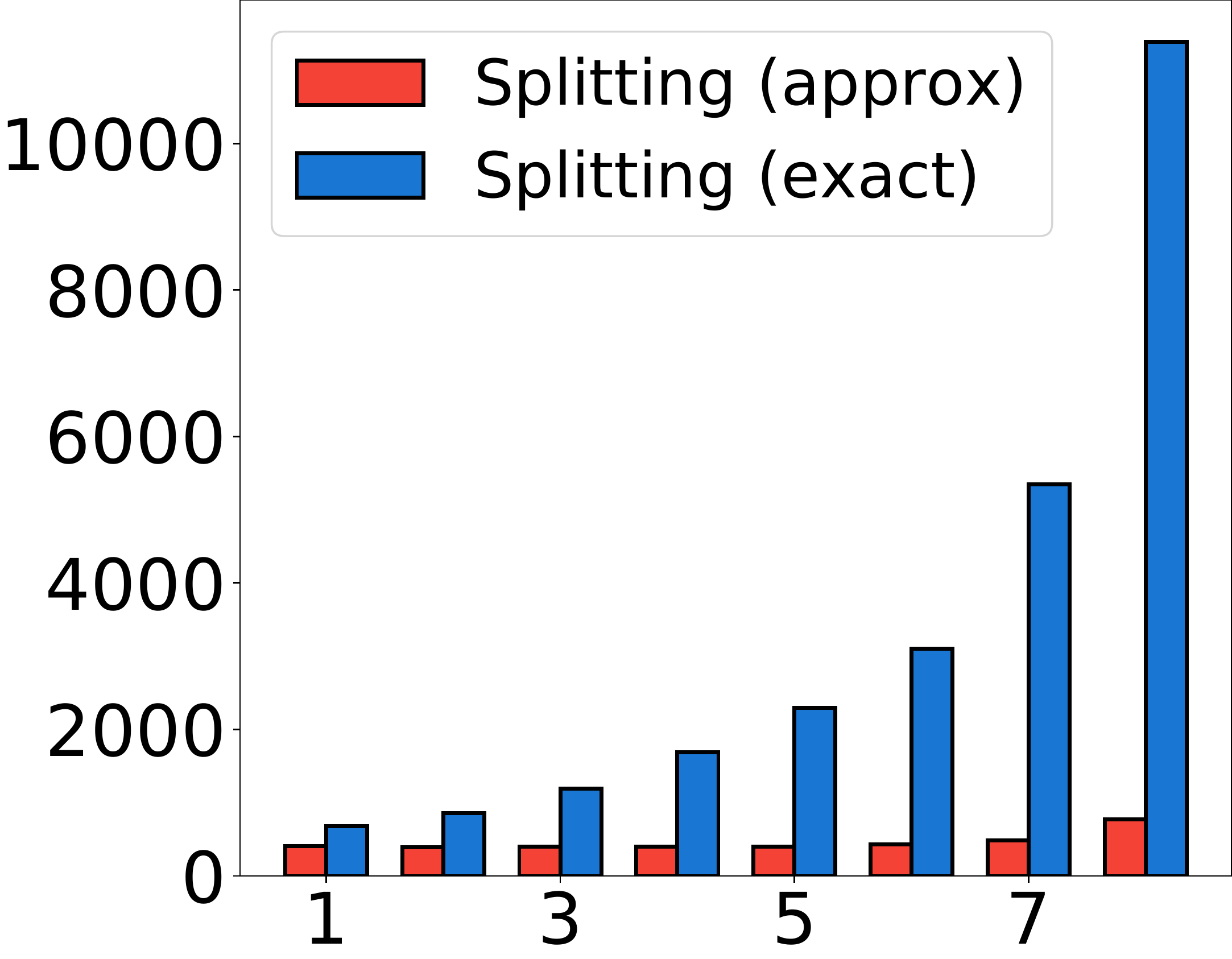} \\
{\scriptsize (a) Log10(flops)} &
{\scriptsize~ (b) ~~\#split}\\
\end{tabular}
\vspace{-1.5em}
\caption{{\small Comparison of testing accuracy (a) and splitting time (b) using exact eigen-decomposition (denoted as \emph{splitting (exact)}) and
our fast gradient-based eigen-approximation (denoted as \emph{splitting (approx)}).}}
\label{fig:mbv1_gradient}
\end{small}
\end{wrapfigure}
In Figure~\ref{fig:mbv1_gradient} (a-b), 
we examine the accuracy and speed of 
our fast gradient-based eigen-approximation. 
We run all methods on a server with one V100 GPU and 16 CPU cores and 
report the wall-clock time. 
We can see that our fast method (red dots and bars) 
achieves almost the same accuracy as the splitting based on exact eigen-decomposition (blue dots and bars), while achieving significant gain in computational time (see Figure~\ref{fig:mbv1_gradient} (b)). 

\subsection{Results on ImageNet}

We conduct experiments on large-scale ImageNet dataset,
on which our method again shows clear advantages over existing methods. 
Note that splitting based on exact eigen-composition is no longer feasible on ImageNet and our fast gradient-based approximation must be used. 

\paragraph{Dataset} 
The ImageNet dataset \citep{deng2009imagenet} consists of about 1.2 million training images, and $50,000$ validation images, classified into $1,000$ distinct classes. We resize the image size to $224\times 224$, and adopt the standard data augment scheme (mirroring and shifting) for training images \citep[e.g.][]{howard2017mobilenets, sandler2018mobilenetv2}.

\paragraph{Settings}
We choose 
both MobileNetV1 \citep{howard2017mobilenets}  and MobileNetV2 \citep{sandler2018mobilenetv2} as our base net for splitting,  
which are strong baselines and  specifically  hand-designed and heavily tuned to optimize accuracy under a flops-constrain on the ImageNet dataset.



For parametric updates, 
we follow standard training settings on the ImageNet dataset using MobileNets.
Specifically,  we train with a batch-size of $128$ on 4 GPUs (total batch size 512).
We use stochastic gradient descent with an initial learning rate $0.2$ and $0.1$ for MobileNetV1 and MobileNetV2, respectively. 
We apply cosine learning rate annealing scheduling and use label smoothing (0.1) 
by following \citep{he2019bag}.

For our method, we start with relative small models (denoted by \emph{Splitting-0 (seed)}) by shrinking the network uniformly with a width multipler 0.3, and gradually grow the network via energy-aware splitting. We use Splitting-$k$ to represent the model we discovered at the $k$-th splitting stage.
We report the single-center-crop validation error of different models.

\paragraph{MobileNetV1 Results} In Table~\ref{tab:imagenet_mbv1}, 
we find that our method achieves about $1\%$ top-1 accuracy improvements in general when targeting similar flops. 
On low-flops regime ($<0.15$G flops), 
our method achieves 3.06\% higher top-1 accuracy compared with MobileNet (0.5X) (with width multiper 0.5).
Also,  the model found by our method is $0.97\%$ and $0.57\%$
higher than 
 prior art pruning methods {AMC} \citep{he2018amc} and MetaPruning \citep{liu2019metapruning}, respectively, 
{when comparing with checkpoints with $\sim\!0.3$G flops}.



\paragraph{MobileNetV2 Results}
From table~\ref{tab:imagenet_mbv2}, 
we find that our splitting models yield better performance compared with prior art expert-designed architectures on all flops-regimes.
Specially, 
out \emph{splitting-3} reaches 72.84 top-1 accuracy;
this yields  an 0.8\% improvement over its corresponding baseline model.
On the low-flops regime, 
our \emph{splitting-2} achieves an 1.96\% top-1 accuracy improvement over MobileNetV2 (0.75x); our \emph{splitting-1} is 1.1\% higher than MobileNetV2 (0.5x).
Our performance is also about 0.9\% higher than AMC when targeting 70\% flops. 

\begin{table}[t]
    \centering
    \begin{tabular}{l|ccc}
        \hline
        Model & MACs (G) & Top-1 Accuracy & Top-5 Accuracy \\
        \hline \hline
        MobileNetV1 (1.0x) & 0.569 & 72.93 & 91.14 \\
        Splitting-4 &  \textbf{0.561} & \textbf{73.96} & \textbf{91.49}  \\
        \hline \hline
        MobileNetV1 (0.75x) & 0.317 & 70.25 & 89.49\\
        AMC {\small \citep{he2018amc}} & 0.301 & 70.50 &	89.30\\
        MetaPruning {\small \citep{liu2019metapruning}} &  0.324 & 70.90 & - \\
        Splitting-3 & \textbf{0.292} & \textbf{71.47} & \textbf{89.67} \\
        \hline \hline
        MobileNetV1 (0.5x) &  0.150 & 65.20 & 86.34\\
        MetaPruning {\small \citep{liu2019metapruning}}  &  0.149 & 66.10 & - \\
        Splitting-2 &  \textbf{0.140} & \textbf{68.26} & \textbf{87.93} \\
         \hline \hline
         Splitting-1  & 0.082 & 64.06 & 85.30 \\
         Splitting-0 (seed) &  0.059 & 59.20 & 81.82 \\ 
         \hline
    \end{tabular}
    \caption{Results of ImageNet classification using MobileNetV1.
    Splitting-$k$ denotes the model we discovered at the $k$-th splitting stage. 
    MAC represents multiply-and-accumulate operations.}
    \label{tab:imagenet_mbv1}
\end{table}

\begin{table}[t]
    \centering
    \begin{tabular}{l|cccc}
        \hline
        Model & MACs (G) & Top-1 Accuracy & Top-5 Accuracy \\
        \hline \hline
        MobileNetV2 (1.0x)  & 0.300 &  72.04 & {90.57} \\
        MetaPruning {\small \citep{liu2019metapruning}}  &  \textbf{0.291} & 72.70 & - \\
        Splitting-3 & {0.298} & \textbf{72.84} & \textbf{90.83} \\
        \hline \hline
        MobileNetV2 (0.75x)  & 0.209 & 69.80 & 89.60 \\
        AMC \citep{he2018amc}  &  0.210 & 70.85 &	{89.91}\\
        Splitting-2 & \textbf{0.208} & \textbf{71.76} & \textbf{90.07} \\
        \hline \hline
        MetaPruning {\small \citep{liu2019metapruning}}  &  0.105 & 65.00 & - \\
        MobileNetV2 (0.5x) & 0.097 & 65.40  & 86.40 \\
        Splitting-1 & \textbf{0.095} & \textbf{66.53} & \textbf{87.00} \\
         \hline \hline
        Splitting-0 (seed) & 0.039 & 55.61 & 79.55 \\
         \hline
    \end{tabular}
    \caption{Results on ImageNet using MobileNetV2.
    Splitting-$k$ denotes the model we discovered at the $k$-th splitting stage. MAC denotes the number of multiply-and-accumulate operations. }
    \label{tab:imagenet_mbv2}
\end{table}

\subsection{Ablation study}

In our algorithm, the growth ratio $\alpha$ controls how many neurons we could split at each splitting stage. In this section, we perform an in-depth analysis of the effect of different $\alpha$ values. 
We also examine the robustness of our splitting method regarding 
randomness  
during the network training and splitting (e.g.  parameters initializations, data shuffle). 

\begin{wrapfigure}{r}{0.6\textwidth}
\begin{small}
\begin{center}
\setlength{\tabcolsep}{1pt}
\vspace{-0.5em}
\begin{tabular}{cc}
\raisebox{2em}{\rotatebox{90}{\scriptsize Test Accuracy}}
\includegraphics[height =0.2\textwidth]{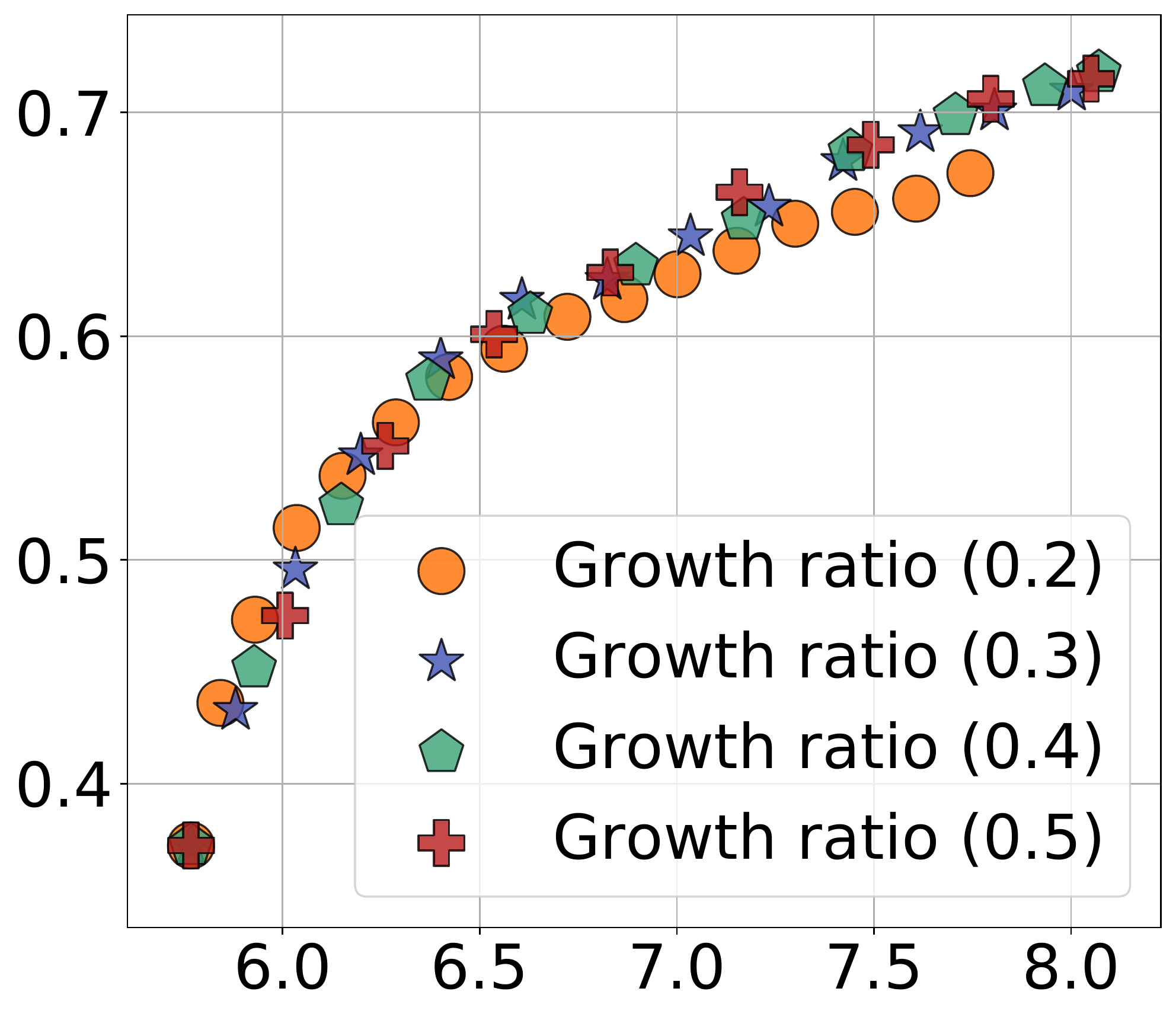} &~~
\includegraphics[height =0.2\textwidth]{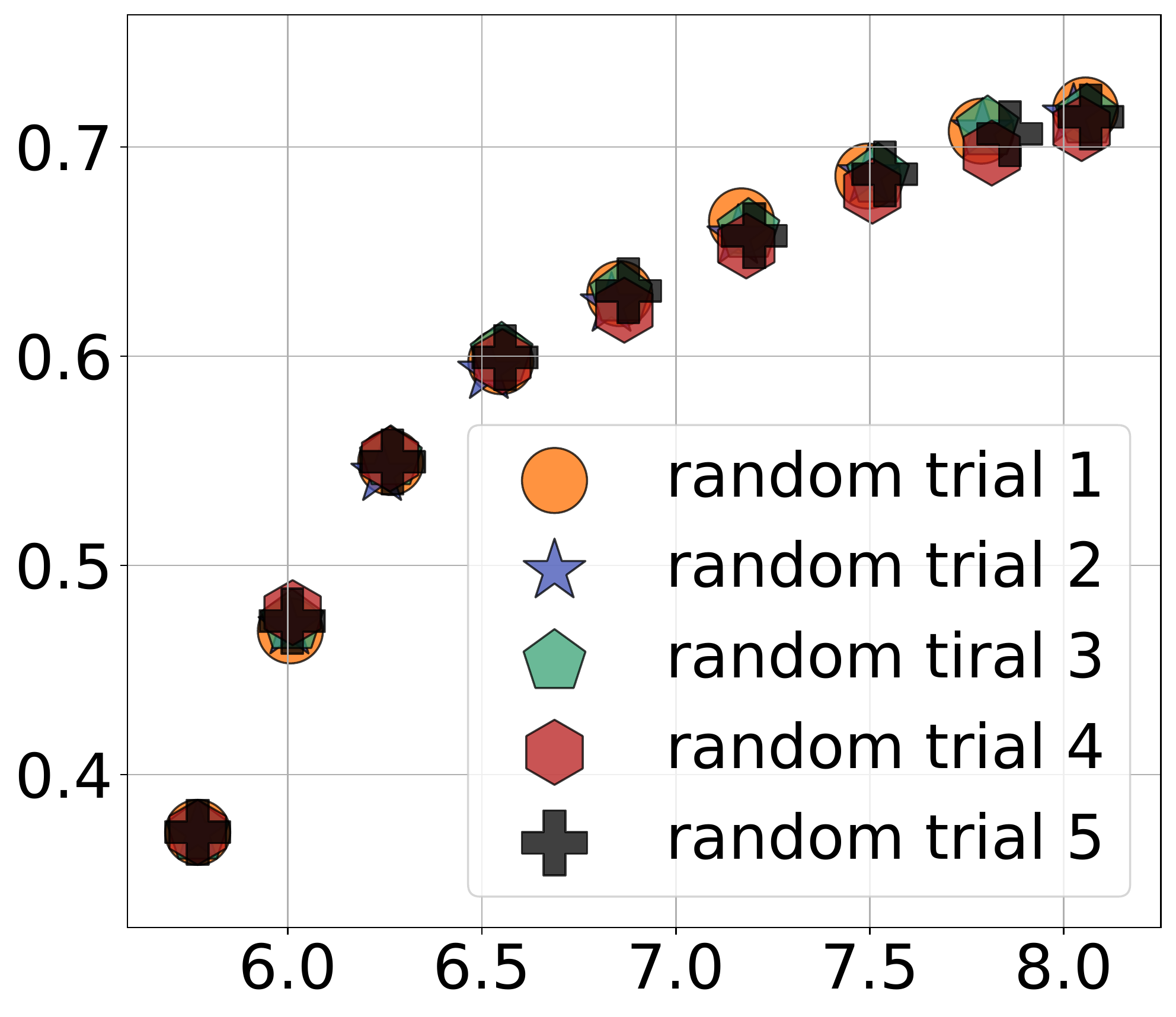} \\
{\scriptsize (a) Log10(flops)} & {\scriptsize (b) Log10(flops)} \\
\end{tabular}
\end{center}
\vspace{-1.5em}
\caption{ Comparison on the test accuracy for various ablation study settings.} 
\label{fig:mbv1_ablation}
\end{small}
\end{wrapfigure}

\paragraph{Impact of growth ratio}  
To find the optimal growth ratio $\alpha$, 
we ran multiple experiments with different growth ratio $\alpha$ under the same settings as section~\ref{sec:cifar100}.
Figure~\ref{fig:mbv1_ablation} (a) shows the performance of various runs. 
We find that the growth ratio in the range of $[0.3, 0.5]$ tend to perform similarly well. 
However, the smaller growth ratio of $\alpha=0.2$ tends to give lower accuracy, 
this may be because with a small growth ratio, the neurons in the layers close to the input may never be selected because of their higher energy cost for splitting,  hence yielding sub-optimal networks. 

\paragraph{Robustness }
We apply our method to grow a small MobileNet \citep{howard2017mobilenets} using different random seeds for parameters initialization and data shuffle under the same setting as Figure~\ref{fig:mbv1_ablation} (a) with a growth ratio 0.5. 
Figure~\ref{fig:mbv1_ablation} (b) shows the test performance
of different models learned.  
As we can see from Figure~\ref{fig:mbv1_ablation} (b), 
all runs perform similarly well with small variations.





\section{Related Work}

Neural architecture search (NAS) has 
been found a powerful tool 
for automating energy-efficient architecture design.
Most existing NAS methods are based on black-box optimization techniques, including reinforcement learning \citep[e.g.][]{zoph2016neural, zoph2018learning} , evolutionary algorithms \citep[e.g.][]{real2019regularized, real2017large}.
However, 
these methods 
are often extremely time-consuming due to the enormous search 
space of possible architectures and 
the high cost for evaluating the performance of each candidate network.  
More recent approaches have made the search more efficient by using weight-sharing \citep[e.g.][]{pham2018efficient, liu2018darts, cai2018proxylessnas},  which, however, 
suffers from the so-called multi-model forgetting problem \citep{benyahia2019overcoming} that causes training instability and performance degradation during search. 
Overall, designing the best architectures using NAS still requires a lot of expert knowledge and trial-and-errors.

In contrast, 
pruning-based methods construct smaller networks from a pretrained over-parameterized neural network by gradually
removing the least important neurons.  
Various pruning strategies 
have been developed based on different heuristics 
\citep[e.g.,][]{han2015deep, li2016pruning, luo2017thinet, he2017channel, peng2019collaborative}, 
including energy-aware pruning methods 
that use energy consumption related metrics to guide the pruning process \citep[e.g.,][]{yang2017designing, gordon2018morphnet, he2018amc, yang2018ecc}. 
{However, a common issue of these methods is to alter 
the standard training objective 
with sparsity-induced regularization which necessities sensitive hyper-parameters tuning.
Furthermore, the final performance is largely limited 
by the initial hand-crafted network, which may not be optimal in the first place.}

\section{Conclusions}
In this work,
we present a fast energy-aware splitting steepest descent approach for 
resource-efficient neural architecture optimization that generalizes 
\citet{splitting2019}.
Empirical results on large-scale ImageNet benchmark using MobileNetV1 and MoibileNetV2 
demonstrate the effectiveness of our method.

\section*{Acknowledgement}
This work is supported in part by NSF CRII 1830161 and NSF CAREER
1846421. 

\bibliography{splitting}

\begin{thebibliography}{3}
\providecommand{\natexlab}[1]{#1}
\providecommand{\url}[1]{\texttt{#1}}
\expandafter\ifx\csname urlstyle\endcsname\relax
  \providecommand{\doi}[1]{doi: #1}\else
  \providecommand{\doi}{doi: \begingroup \urlstyle{rm}\Url}\fi

\bibitem[Bengio \& LeCun(2007)Bengio and LeCun]{Bengio+chapter2007}
Yoshua Bengio and Yann LeCun.
\newblock Scaling learning algorithms towards {AI}.
\newblock In \emph{Large Scale Kernel Machines}. MIT Press, 2007.

\bibitem[Goodfellow et~al.(2016)Goodfellow, Bengio, Courville, and
  Bengio]{goodfellow2016deep}
Ian Goodfellow, Yoshua Bengio, Aaron Courville, and Yoshua Bengio.
\newblock \emph{Deep learning}, volume~1.
\newblock MIT Press, 2016.

\bibitem[Hinton et~al.(2006)Hinton, Osindero, and Teh]{Hinton06}
Geoffrey~E. Hinton, Simon Osindero, and Yee~Whye Teh.
\newblock A fast learning algorithm for deep belief nets.
\newblock \emph{Neural Computation}, 18:\penalty0 1527--1554, 2006.

\end{thebibliography}


\begin{thebibliography}{35}
\providecommand{\natexlab}[1]{#1}
\providecommand{\url}[1]{\texttt{#1}}
\expandafter\ifx\csname urlstyle\endcsname\relax
  \providecommand{\doi}[1]{doi: #1}\else
  \providecommand{\doi}{doi: \begingroup \urlstyle{rm}\Url}\fi

\bibitem[Benyahia et~al.(2019)Benyahia, Yu, Bennani-Smires, Jaggi, Davison,
  Salzmann, and Musat]{benyahia2019overcoming}
Yassine Benyahia, Kaicheng Yu, Kamil Bennani-Smires, Martin Jaggi, Anthony
  Davison, Mathieu Salzmann, and Claudiu Musat.
\newblock Overcoming multi-model forgetting.
\newblock In \emph{Proceedings of the 36th International Conference on Machine
  Learning (ICML)}, 2019.

\bibitem[Cai et~al.(2019)Cai, Zhu, and Han]{cai2018proxylessnas}
Han Cai, Ligeng Zhu, and Song Han.
\newblock Proxyless{NAS}: Direct neural architecture search on target task and
  hardware.
\newblock In \emph{International Conference on Learning Representations
  (ICLR)}, 2019.

\bibitem[Dantzig(1998)]{dantzig1998linear}
George~Bernard Dantzig.
\newblock \emph{Linear programming and extensions}.
\newblock Princeton university press, 1998.

\bibitem[Deng et~al.(2009)Deng, Dong, Socher, Li, Li, and
  Fei-Fei]{deng2009imagenet}
Jia Deng, Wei Dong, Richard Socher, Li-Jia Li, Kai Li, and Li~Fei-Fei.
\newblock Imagenet: A large-scale hierarchical image database.
\newblock In \emph{2009 IEEE conference on computer vision and pattern
  recognition (CVPR)}, pp.\  248--255. IEEE, 2009.

\bibitem[Devlin et~al.(2018)Devlin, Chang, Lee, and Toutanova]{devlin2018bert}
Jacob Devlin, Ming-Wei Chang, Kenton Lee, and Kristina Toutanova.
\newblock Bert: Pre-training of deep bidirectional transformers for language
  understanding.
\newblock In \emph{Annual Conference of the North American Chapter of the
  Association for Computational Linguistics: Human Language Technologies
  (NAACL-HLT)}, 2018.

\bibitem[Gordon et~al.(2018)Gordon, Eban, Nachum, Chen, Wu, Yang, and
  Choi]{gordon2018morphnet}
Ariel Gordon, Elad Eban, Ofir Nachum, Bo~Chen, Hao Wu, Tien-Ju Yang, and Edward
  Choi.
\newblock Morphnet: Fast \& simple resource-constrained structure learning of
  deep networks.
\newblock In \emph{Proceedings of the IEEE Conference on Computer Vision and
  Pattern Recognition (CVPR)}, pp.\  1586--1595, 2018.

\bibitem[Han et~al.(2016)Han, Mao, and Dally]{han2015deep}
Song Han, Huizi Mao, and William~J Dally.
\newblock Deep compression: Compressing deep neural networks with pruning,
  trained quantization and huffman coding.
\newblock \emph{International Conference on Learning Representations (ICLR)},
  2016.

\bibitem[He et~al.(2016)He, Zhang, Ren, and Sun]{he2016deep}
Kaiming He, Xiangyu Zhang, Shaoqing Ren, and Jian Sun.
\newblock Deep residual learning for image recognition.
\newblock In \emph{Proceedings of the IEEE conference on computer vision and
  pattern recognition (CVPR)}, pp.\  770--778, 2016.

\bibitem[He et~al.(2017{\natexlab{a}})He, Gkioxari, Doll{\'a}r, and
  Girshick]{he2017mask}
Kaiming He, Georgia Gkioxari, Piotr Doll{\'a}r, and Ross Girshick.
\newblock Mask r-cnn.
\newblock In \emph{Proceedings of the IEEE international conference on computer
  vision and pattern recognition (CVPR)}, pp.\  2961--2969, 2017{\natexlab{a}}.

\bibitem[He et~al.(2019)He, Zhang, Zhang, Zhang, Xie, and Li]{he2019bag}
Tong He, Zhi Zhang, Hang Zhang, Zhongyue Zhang, Junyuan Xie, and Mu~Li.
\newblock Bag of tricks for image classification with convolutional neural
  networks.
\newblock In \emph{Proceedings of the IEEE Conference on Computer Vision and
  Pattern Recognition (CVPR)}, pp.\  558--567, 2019.

\bibitem[He et~al.(2017{\natexlab{b}})He, Zhang, and Sun]{he2017channel}
Yihui He, Xiangyu Zhang, and Jian Sun.
\newblock Channel pruning for accelerating very deep neural networks.
\newblock In \emph{Proceedings of the IEEE International Conference on Computer
  Vision (ICCV)}, pp.\  1389--1397, 2017{\natexlab{b}}.

\bibitem[He et~al.(2018)He, Lin, Liu, Wang, Li, and Han]{he2018amc}
Yihui He, Ji~Lin, Zhijian Liu, Hanrui Wang, Li-Jia Li, and Song Han.
\newblock Amc: Automl for model compression and acceleration on mobile devices.
\newblock In \emph{Proceedings of the European Conference on Computer Vision
  (ECCV)}, pp.\  784--800, 2018.

\bibitem[Howard et~al.(2017)Howard, Zhu, Chen, Kalenichenko, Wang, Weyand,
  Andreetto, and Adam]{howard2017mobilenets}
Andrew~G Howard, Menglong Zhu, Bo~Chen, Dmitry Kalenichenko, Weijun Wang,
  Tobias Weyand, Marco Andreetto, and Hartwig Adam.
\newblock Mobilenets: Efficient convolutional neural networks for mobile vision
  applications.
\newblock \emph{arXiv preprint arXiv:1704.04861}, 2017.

\bibitem[Huang et~al.(2017)Huang, Liu, Van Der~Maaten, and
  Weinberger]{huang2017densely}
Gao Huang, Zhuang Liu, Laurens Van Der~Maaten, and Kilian~Q Weinberger.
\newblock Densely connected convolutional networks.
\newblock In \emph{Proceedings of the IEEE conference on computer vision and
  pattern recognition (CVPR)}, pp.\  4700--4708, 2017.

\bibitem[Li et~al.(2017)Li, Kadav, Durdanovic, Samet, and Graf]{li2016pruning}
Hao Li, Asim Kadav, Igor Durdanovic, Hanan Samet, and Hans~Peter Graf.
\newblock Pruning filters for efficient convnets.
\newblock \emph{International Conference on Learning Representations (ICLR)},
  2017.

\bibitem[Liu et~al.(2019{\natexlab{a}})Liu, Simonyan, and Yang]{liu2018darts}
Hanxiao Liu, Karen Simonyan, and Yiming Yang.
\newblock {DARTS}: Differentiable architecture search.
\newblock \emph{International Conference on Learning Representations (ICLR)},
  2019{\natexlab{a}}.

\bibitem[Liu et~al.(2019{\natexlab{b}})Liu, Wu, and Wang]{splitting2019}
Qiang Liu, Lemeng Wu, and Dilin Wang.
\newblock Splitting steepest descent for growing neural architectures.
\newblock \emph{Advances in Neural Information Processing Systems}, pp.\
  10656--10666, 2019{\natexlab{b}}.

\bibitem[Liu et~al.(2019{\natexlab{c}})Liu, Mu, Zhang, Guo, Yang, Cheng, and
  Sun]{liu2019metapruning}
Zechun Liu, Haoyuan Mu, Xiangyu Zhang, Zichao Guo, Xin Yang, Tim Kwang-Ting
  Cheng, and Jian Sun.
\newblock Metapruning: Meta learning for automatic neural network channel
  pruning.
\newblock \emph{arXiv preprint arXiv:1903.10258}, 2019{\natexlab{c}}.

\bibitem[Liu et~al.(2017)Liu, Li, Shen, Huang, Yan, and Zhang]{liu2017learning}
Zhuang Liu, Jianguo Li, Zhiqiang Shen, Gao Huang, Shoumeng Yan, and Changshui
  Zhang.
\newblock Learning efficient convolutional networks through network slimming.
\newblock In \emph{Proceedings of the IEEE International Conference on Computer
  Vision (ICCV)}, pp.\  2736--2744, 2017.

\bibitem[Liu et~al.(2019{\natexlab{d}})Liu, Sun, Zhou, Huang, and
  Darrell]{liu2018rethinking}
Zhuang Liu, Mingjie Sun, Tinghui Zhou, Gao Huang, and Trevor Darrell.
\newblock Rethinking the value of network pruning.
\newblock \emph{International Conference on Learning Representations (ICLR)},
  2019{\natexlab{d}}.

\bibitem[Luo et~al.(2017)Luo, Wu, and Lin]{luo2017thinet}
Jian-Hao Luo, Jianxin Wu, and Weiyao Lin.
\newblock Thinet: A filter level pruning method for deep neural network
  compression.
\newblock In \emph{Proceedings of the IEEE International Conference on Computer
  Vision (ICCV)}, pp.\  5058--5066, 2017.

\bibitem[Parlett(1998)]{parlett1998symmetric}
Beresford~N Parlett.
\newblock \emph{The symmetric eigenvalue problem}, volume~20.
\newblock siam, 1998.

\bibitem[Paszke et~al.(2017)Paszke, Gross, Chintala, Chanan, Yang, DeVito, Lin,
  Desmaison, Antiga, and Lerer]{paszke2017automatic}
Adam Paszke, Sam Gross, Soumith Chintala, Gregory Chanan, Edward Yang, Zachary
  DeVito, Zeming Lin, Alban Desmaison, Luca Antiga, and Adam Lerer.
\newblock Automatic differentiation in pytorch.
\newblock 2017.

\bibitem[Peng et~al.(2019)Peng, Wu, Chen, and Huang]{peng2019collaborative}
Hanyu Peng, Jiaxiang Wu, Shifeng Chen, and Junzhou Huang.
\newblock Collaborative channel pruning for deep networks.
\newblock In \emph{International Conference on Machine Learning (ICML)}, pp.\
  5113--5122, 2019.

\bibitem[Pham et~al.(2018)Pham, Guan, Zoph, Le, and Dean]{pham2018efficient}
Hieu Pham, Melody~Y Guan, Barret Zoph, Quoc~V Le, and Jeff Dean.
\newblock Efficient neural architecture search via parameter sharing.
\newblock In \emph{Proceedings of the 35th International Conference on Machine
  Learning (ICML)}, 2018.

\bibitem[Real et~al.(2017)Real, Moore, Selle, Saxena, Suematsu, Tan, Le, and
  Kurakin]{real2017large}
Esteban Real, Sherry Moore, Andrew Selle, Saurabh Saxena, Yutaka~Leon Suematsu,
  Jie Tan, Quoc~V Le, and Alexey Kurakin.
\newblock Large-scale evolution of image classifiers.
\newblock 2017.

\bibitem[Real et~al.(2019)Real, Aggarwal, Huang, and Le]{real2019regularized}
Esteban Real, Alok Aggarwal, Yanping Huang, and Quoc~V Le.
\newblock Regularized evolution for image classifier architecture search.
\newblock In \emph{Proceedings of the AAAI Conference on Artificial
  Intelligence (AAAI)}, volume~33, pp.\  4780--4789, 2019.

\bibitem[Rumelhart et~al.(1988)Rumelhart, Hinton, Williams,
  et~al.]{rumelhart1988learning}
David~E Rumelhart, Geoffrey~E Hinton, Ronald~J Williams, et~al.
\newblock Learning representations by back-propagating errors.
\newblock \emph{Cognitive modeling}, 5\penalty0 (3):\penalty0 1, 1988.

\bibitem[Sandler et~al.(2018)Sandler, Howard, Zhu, Zhmoginov, and
  Chen]{sandler2018mobilenetv2}
Mark Sandler, Andrew Howard, Menglong Zhu, Andrey Zhmoginov, and Liang-Chieh
  Chen.
\newblock Mobilenetv2: Inverted residuals and linear bottlenecks.
\newblock In \emph{Proceedings of the IEEE Conference on Computer Vision and
  Pattern Recognition (CVPR)}, pp.\  4510--4520, 2018.

\bibitem[Simonyan \& Zisserman(2015)Simonyan and Zisserman]{simonyan2014very}
Karen Simonyan and Andrew Zisserman.
\newblock Very deep convolutional networks for large-scale image recognition.
\newblock In \emph{International Conference on Learning Representations
  (ICLR)}, 2015.

\bibitem[Tieleman \& Hinton(2012)Tieleman and Hinton]{tieleman2012lecture}
Tijmen Tieleman and Geoffrey Hinton.
\newblock Lecture 6.5-rmsprop: Divide the gradient by a running average of its
  recent magnitude.
\newblock \emph{COURSERA: Neural networks for machine learning}, 4\penalty0
  (2):\penalty0 26--31, 2012.

\bibitem[Yang et~al.(2019)Yang, Zhu, and Liu]{yang2018ecc}
Haichuan Yang, Yuhao Zhu, and Ji~Liu.
\newblock Ecc: Platform-independent energy-constrained deep neural network
  compression via a bilinear regression model.
\newblock In \emph{The IEEE Conference on Computer Vision and Pattern
  Recognition (CVPR)}, June 2019.

\bibitem[Yang et~al.(2017)Yang, Chen, and Sze]{yang2017designing}
Tien-Ju Yang, Yu-Hsin Chen, and Vivienne Sze.
\newblock Designing energy-efficient convolutional neural networks using
  energy-aware pruning.
\newblock In \emph{Proceedings of the IEEE Conference on Computer Vision and
  Pattern Recognition (CVPR)}, pp.\  5687--5695, 2017.

\bibitem[Zoph \& Le(2017)Zoph and Le]{zoph2016neural}
Barret Zoph and Quoc~V Le.
\newblock Neural architecture search with reinforcement learning.
\newblock \emph{Proceedings of the 35th International Conference on Machine
  Learning (ICML)}, 2017.

\bibitem[Zoph et~al.(2018)Zoph, Vasudevan, Shlens, and Le]{zoph2018learning}
Barret Zoph, Vijay Vasudevan, Jonathon Shlens, and Quoc~V Le.
\newblock Learning transferable architectures for scalable image recognition.
\newblock In \emph{Proceedings of the IEEE conference on computer vision and
  pattern recognition (CVPR)}, pp.\  8697--8710, 2018.

\end{thebibliography}
\bibliographystyle{iclr2020_conference}


\end{document}